\documentclass[conference,a4paper]{IEEEtran}

\usepackage[a4paper,
            left=0.63in,
            right=0.63in,
            top=0.75in,
            bottom=1in,
            headsep=0.5in,
            footskip=0.5in,
            columnsep=0.25in]{geometry}
\IEEEoverridecommandlockouts
\usepackage{cite}
\usepackage{fancyhdr}
\usepackage{amsmath,amssymb,amsfonts}
\usepackage{algorithmic}
\usepackage{graphicx}
\usepackage{textcomp}
\usepackage{xcolor}
\def\BibTeX{{\rm B\kern-.05em{\sc i\kern-.025em b}\kern-.08em
    T\kern-.1667em\lower.7ex\hbox{E}\kern-.125emX}}

\begin{document}

\newgeometry{left=0.62in,
            right=0.62in,
            top=0.38in,
            bottom=1in,
            headsep=0.5in,
            footskip=0.5in,
            columnsep=0.25in}

\title{Script Sensitivity: Benchmarking Language Models on Unicode, Romanized and Mixed-Script Sinhala}

\author{\IEEEauthorblockN{Minuri Rajapakse}
\IEEEauthorblockA{\textit{School of Computing} \\
\textit{Informatics Institute of Technology}\\
Colombo, Sri Lanka \\
minuri.20220646@iit.ac.lk}
\and
\IEEEauthorblockN{Ruvan Weerasinghe}
\IEEEauthorblockA{\textit{School of Computing} \\
\textit{Informatics Institute of Technology}\\
Colombo, Sri Lanka \\
ruvan.w@iit.ac.lk}}


\maketitle

\begin{abstract}
The performance of Language Models (LMs) on low-resource, morphologically rich languages like Sinhala remains largely unexplored, particularly regarding script variation in digital communication. Sinhala exhibits script duality, with Unicode used in formal contexts and Romanized text dominating social media, while mixed-script usage is common in practice. This paper benchmarks 24 open-source LMs on Unicode, Romanized and mixed-script Sinhala using perplexity evaluation across diverse text sources. Results reveal substantial script sensitivity, with median performance degradation exceeding 300 times from Unicode to Romanized text. Critically, model size shows no correlation with script-handling competence, as smaller models often outperform architectures 28 times larger. Unicode performance strongly predicts mixed-script robustness but not Romanized capability, demonstrating that single-script evaluation substantially underestimates real-world deployment challenges. These findings establish baseline LM capabilities for Sinhala and provide practical guidance for model selection in multi-script low-resource environments.
\end{abstract}


\section{Introduction}
Language Models (LMs) have demonstrated remarkable capabilities across a wide range of Natural Language Processing (NLP) tasks. However, their development and evaluation have been overwhelmingly centered on high-resource languages such as English, leaving many low-resource and morphologically rich languages underexplored. Sinhala, an Indo-Aryan language spoken by about 22 million people, exemplifies this gap due to its linguistic complexity and limited representation in large-scale training corpora\cite{sitse}.

Sinhala poses unique challenges for modern NLP systems. Sinhala is morphologically rich and agglutinative, producing many word forms from a single root, and exhibits pronounced script variation in digital communication. While Unicode Sinhala is predominantly used in formal, educational and literary contexts, Romanized Sinhala, written using the Latin alphabet, dominates informal platforms such as social media, messaging applications and online forums. In practice, users frequently combine both scripts within the same sentence or conversation, resulting in mixed-script Sinhala text. Romanized Sinhala lacks standardization, creating a gap between Unicode-focused benchmarks and real-world Romanized or mixed-script use.

Recent benchmarking efforts for Language Models have emphasized downstream task performance or multilingual evaluation across a broad set of languages. In such benchmarks, Sinhala is either minimally represented or evaluated indirectly through translation or classification tasks\cite{sinhalammlu}. As a result, there is limited understanding of how contemporary LMs perform on Sinhala intrinsically and how their capabilities vary across Unicode, Romanized and mixed-script forms. Despite widespread use of Romanized Sinhala in digital communication, no systematic evaluation exists comparing LM performance across Unicode, Romanized and mixed-script conditions.

To address these gaps, this paper presents a systematic benchmark of 24 modern open-source generative language models (ranging from 350M to 14B parameters) on Unicode, Romanized and mixed-script Sinhala text. Perplexity as an intrinsic evaluation metric is used to assess core language modeling ability across 1,500 sentences (500 per script condition). Perplexity provides insight into how well a model captures the probability distribution of a language, independent of any specific downstream task. It is well-suited for evaluating script variation because it is sensitive to token-level modeling differences caused by script-dependent representations. A mixed-script evaluation setting is also introduced to measure robustness under realistic script-switching conditions.

Specifically, this work addresses three questions: (1) how does LM performance vary across Unicode, Romanized and mixed-script Sinhala; (2) does strong Unicode performance predict capability on Romanized or mixed-script text; and (3) does model scale correlate with script-handling competence in a low-resource, multi-script setting.

\section{Related Work}

\subsection{Sinhala NLP and Low-Resource Language Modeling}

Sinhala is a morphologically rich and agglutinative language with limited high-quality digital resources, posing substantial challenges for modern NLP systems. Prior research has focused on task-specific applications, such as neural machine translation and transliteration. Synthetic data augmentation addressed rare-word problems in Sinhala Tamil translation\cite{tennage}, while Transformer-based architectures with mBART fine-tuning achieved gains for Sinhala English tasks\cite{thillainathan2021finetuning}. Recent transliteration models bridge Romanized and Unicode Sinhala to improve translation accuracy\cite{demel2025sinhala}. 

While these studies contribute valuable datasets and evaluation protocols, they do not assess the foundational language modeling capability of modern generative models on Sinhala text itself. Script variation is common in many low-resource languages. Hindi (Hinglish) \cite{dey2024social} and Arabic (Arabizi) \cite{badrashinyby201arabizi} frequently appear in Romanized form on social media, reflecting broader multilingual script challenges. Prior work shows that Romanization introduces token fragmentation\cite{madhani2023improving}\cite{ahia2023tail}\cite{petrov2023language} and degraded model performance \cite{chakrabarty2024effect}. 
\restoregeometry
Despite Romanized Sinhala's prevalence in informal communication, systematic evaluations comparing Unicode and Romanized Sinhala are virtually absent. As illustrated in Table~\ref{tab:romanization_examples}, a single Unicode word can have multiple Romanized representations, reflecting user preferences and a lack of standardization. This variability poses significant challenges for language model evaluation.

\begin{table}[htbp]
\centering
\footnotesize
\caption{Romanization Variability in Sinhala}
\label{tab:romanization_examples}
\renewcommand{\arraystretch}{1.5}
\begin{tabular}{|c|l|}
\hline
\textbf{Unicode} & \textbf{Romanized Variants} \\
\hline
\includegraphics[height=0.3cm]{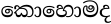} & kohomada, kohomd, kohomde \\
\hline
\includegraphics[height=0.35cm]{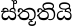} & sthutiyi, stutiyi, sthuthi \\
\hline
\includegraphics[height=0.3cm]{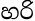} & hari, hri, harri \\
\hline
\end{tabular}
\end{table}

Existing work has made valuable contributions to transliteration\cite{roark2020processing} and script conversion\cite{khullar2024script}\cite{j2024romansetu}, but systematic evaluation of language model performance on Romanized and mixed-script variants as independent script conditions is still needed.

Multilingual benchmarks such as GLUE\cite{wang2018glue}, XTREME\cite{hu2020xtreme} and MMLU\cite{hendrycks2021measuring} predominantly emphasize high-resource languages or downstream tasks, 
leaving low-resource languages underrepresented\cite{ranathunga2022some}. Some include Sinhala \cite{sinhalammlu}, but typically as a minor subset, overlooking informal Romanized variants that dominate real-world digital communication\cite{khullar2024script}\cite{j2024romansetu}. Perplexity-based benchmarking for Sinhala and South Asian languages remains extremely limited in the context of modern generative LMs\cite{leong2025seahelm}.

\subsection{Research Gap and Contribution}

Four critical gaps emerge in the Sinhala NLP field. First, modern LMs lack intrinsic benchmarking on Sinhala using foundational metrics like perplexity, with existing work focusing on downstream tasks rather than core language modeling competence. Second, despite widespread parallel use of Unicode and Romanized Sinhala, no comparative analysis systematically evaluates model performance across both scripts.

Third, mixed-script Sinhala, where Unicode and Romanized scripts co-occur within the same text, is pervasive on social media and messaging platforms, yet LM robustness under such realistic mixed-script conditions remains largely unexplored. Fourth, there is a limited understanding of how tokenizer design and pre-training data composition influence script-dependent performance divergence, particularly whether strong single-script performance translates to robustness under script mixing.

This benchmark directly addresses all four gaps through systematic intrinsic evaluation across three script conditions. By releasing the evaluation datasets publicly, this work aims to establish a replicable benchmarking protocol that the research community can directly adopt for evaluating LMs on other script-diverse low-resource languages without requiring new data collection.


\section{Methodology}

\begin{figure}[h]
\centering
\includegraphics[width=\columnwidth]{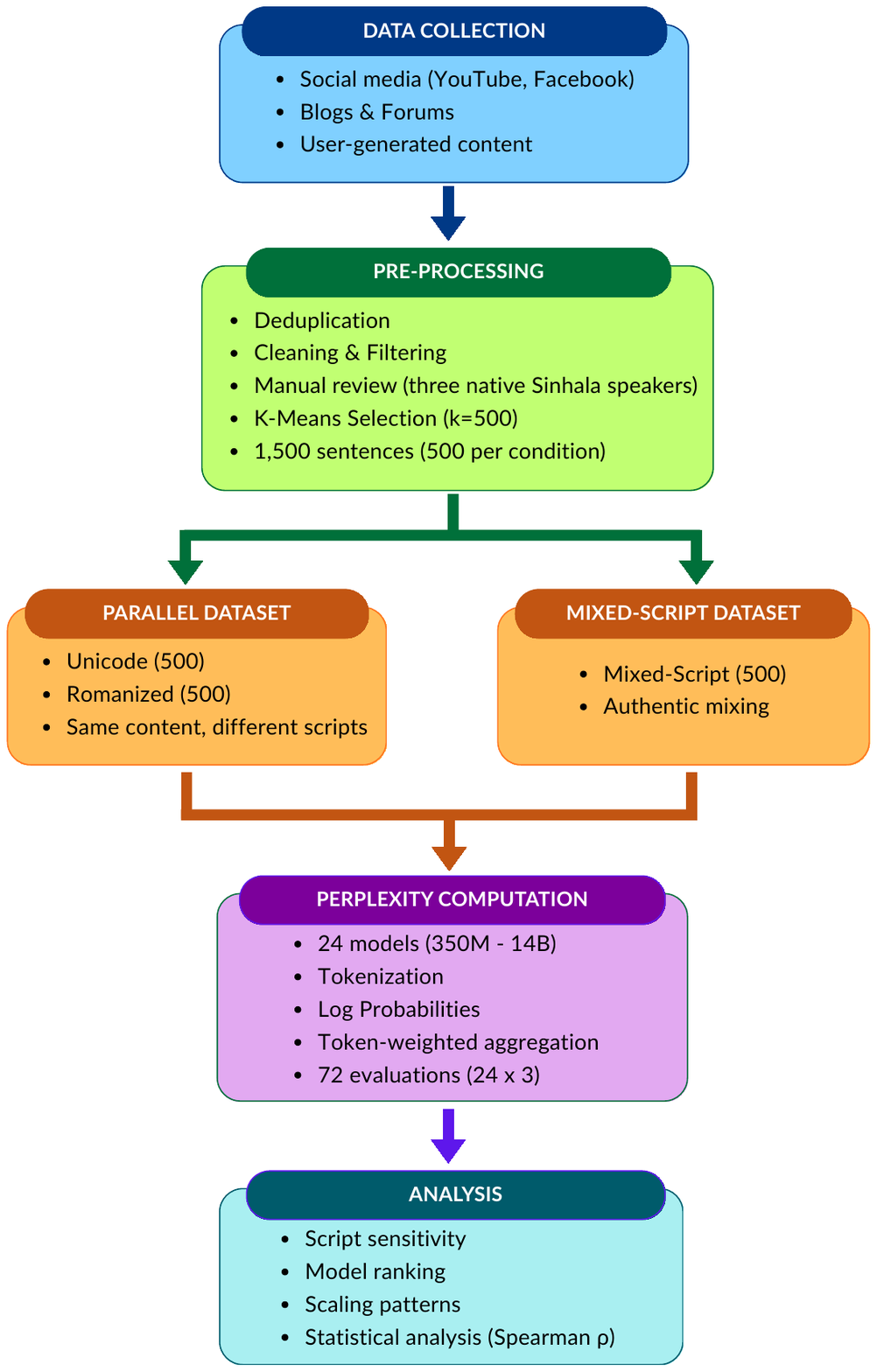}
\caption{Benchmark methodology workflow from data collection through perplexity evaluation.}
\label{workflowfig}
\end{figure}

This section describes the approach to benchmarking Language Models on Unicode, Romanized and mixed-script Sinhala text. Fig.\ref{workflowfig} illustrates the complete workflow from data collection through perplexity evaluation.

\subsection{Dataset Construction}

A novel Sinhala evaluation corpus was constructed to capture realistic script variation patterns in Sri Lankan digital communication. The corpus consists of two complementary datasets: a parallel Unicode-Romanized dataset\cite{uniromandataset} and a separate mixed-script dataset\cite{codemixdataset}, each addressing distinct evaluation needs.

Sinhala text was collected from four authentic sources: YouTube comments, Facebook posts, blogs covering technology and culture, and online forum discussions, ensuring diversity in register and domain. Data collection focused exclusively on publicly available content and followed ethical guidelines, excluding personally identifiable information.

Raw text underwent systematic preprocessing to ensure quality and consistency. Preprocessing removed URLs, email addresses, hashtags and non-textual elements, filtered non-Sinhala content, and applied deduplication to eliminate repeated posts and bot-generated content. Sentence segmentation was performed with length filtering applied to retain sentences between 2 and 15 tokens. All text was manually reviewed and curated.

\subsection{Evaluation Set Design}

Two distinct evaluation datasets from the cleaned corpus were constructed, each serving a specific benchmarking purpose.

\subsubsection{Unicode-Romanized Parallel Dataset}
This dataset contains 500 sentence pairs with identical semantic content in both Unicode Sinhala and Romanized transliteration. This parallel structure enables direct comparison of model performance across script conditions, isolating script representation effects from semantic complexity.

\subsubsection{Mixed-Script Dataset} 
The second dataset comprises 500 sentences of authentic mixed-script Sinhala, where Unicode and Romanized scripts co-occur within individual sentences. These sentences preserve natural code-switching patterns from user-generated content, reflecting the dominant communication style in Sri Lankan social media.

To maximize linguistic diversity while avoiding redundancy, a clustering-based selection strategy was employed rather than random sampling for both datasets. For each script condition, sentence embeddings were computed using the sentence-transformers/paraphrase-multilingual-MiniLM-L12-v2 model, a widely used multilingual encoder trained to produce semantically aligned representations across more than 50 languages\cite{sentBert}. K-means clustering with k=500 was applied to partition sentences into semantic clusters based on their embedding representations. From each cluster, the sentence nearest to the centroid was selected as the representative example. This approach ensures that evaluation sets span diverse topics and linguistic patterns while minimizing within-cluster redundancy. 

The final evaluation corpus comprises 1,500 sentences total across both datasets. The Unicode-Romanized dataset contains 500 sentence pairs. Both Unicode and Romanized subsets share identical mean length (6.07 ± 1.83 tokens), with vocabulary sizes of 1,184 and 1,171, respectively. The mixed-script dataset contains 500 sentences with a mean length of 9.49 ± 4.03 tokens and a vocabulary size of 2,262. Unlike synthetically transliterated datasets, the mixed-script corpus preserves authentic patterns including lexical borrowing, code-switching, and inconsistent spelling.

\subsection{Model Selection}

This work evaluates 24 open-source generative language models spanning multiple architectures, parameter scales and training paradigms. Models were selected based on three criteria: (1) availability through the Hugging Face Model Hub, (2) adherence to decoder-only transformer architectures, suitable for generative language modeling and (3) diversity in scale, with the parameter counts ranging from 350M to 14B to assess scaling effects on script-handling capability.

Decoder-only architectures were selected as they represent the dominant paradigm in modern generative language modeling, enabling direct comparison across 
the widest range of contemporary open-source models.

The evaluation suite spans 350M to 14B parameters across five size categories: three tiny models (Pythia-410M, Opt-350M, Bloom-560M), seven small (1-2B), five medium (2.7-3B), seven large (7-8B including Llama 3.1-8B, Mistral-7B, Qwen2-7B) and two very large models (Mistral-Nemo-12B, Phi-4-14B). This distribution enables analysis of both within-family scaling behavior, such as comparing Llama 3.2-1B through Llama 3.1-8B, and cross-family architectural differences across model lineages.

\subsection{Evaluation Metrics}

Perplexity is used here as the primary intrinsic metric for assessing language modeling quality. Perplexity measures how well a probability model predicts a sequence. For a test sequence of N tokens $w_1, w_2, ..., w_N$, perplexity is computed as the exponentiated average negative log-likelihood:

\begin{equation}
\label{eq:perplexity}
\mathrm{PPL}(W) = \exp\left( -\frac{1}{N} \sum_{i=1}^{N} \log P(w_i \mid w_1, \ldots, w_{i-1}) \right)
\end{equation}

Lower perplexity indicates better predictive performance, reflecting 
that the model assigns a higher probability to observed sequences. This 
metric enables direct comparison of language modeling quality across 
diverse architectures and training regimes.

Perplexity is computed separately for each script condition (Unicode, Romanized and mixed-script), enabling direct comparison of model robustness across script variations. Cross-script performance divergence is quantified through perplexity ratios and rank correlation analysis to assess whether models that excel on Unicode Sinhala maintain comparable performance on Romanized and mixed-script inputs.

\subsection{Experimental Setup}

All experiments were conducted on NVIDIA A100 GPUs with 40GB memory using PyTorch 2.0 and Hugging Face Transformers library version 4.36. Models were loaded in float16 precision to balance computational efficiency and numerical stability. For each model and sentence pair, the negative log-likelihood of the sentence was computed for the given model, summing across all tokens and exponentiating to obtain the perplexity score. Tokenization was performed using each model's native tokenizer without modification to preserve authentic tokenization behavior and its impact on script handling.

For each model, perplexity was computed over the entire evaluation set by aggregating token-level log-likelihoods across all 500 sentences per script condition and exponentiating the average negative log-likelihood. This methodology enables direct comparison of model performance across Unicode, Romanized and mixed-script Sinhala while isolating the effects of script variation on language modeling quality.

\section{Results}

This section presents perplexity results for 24 open-source language models evaluated on Unicode, Romanized and mixed-script Sinhala.

\subsection{Overall Perplexity Across Scripts}

Table \ref{ppt_table} summarizes model perplexities on Unicode, Romanized and mixed-script Sinhala. Perplexity values confirm that all models find Sinhala substantially harder than high-resource languages, with large variation across architectures and script types.

Models generally achieve the lowest perplexity on Unicode Sinhala, higher perplexity on Romanized Sinhala and intermediate values on mixed-script text. In contrast, many models show Romanized perplexities above 1,000 with some exceeding 3,000, indicating that Romanized Sinhala remains poorly modeled despite its prevalence in informal communication.

Mixed-script perplexities are consistently higher than Unicode perplexities but significantly lower than Romanized perplexities for most models, suggesting that the presence of Unicode tokens partially anchors the model’s predictions even when Romanized segments are poorly handled. This pattern highlights the sensitivity of current LMs to script choice and suggests that tokenizer design and training data composition play a central role in script robustness.

\begin{table}[htbp]
\caption{Perplexity of Open-Source Models on Unicode, Romanized and Mixed-Script Sinhala}
\centering
\footnotesize
\setlength{\tabcolsep}{3pt}
\renewcommand{\arraystretch}{1.2}
\begin{tabular}{|l|l|l|l|l|l|}
\hline
\textbf{Model} & \textbf{Params} & \textbf{Unicode} & \textbf{Romanized} & \textbf{Mixed} \\
\hline
Pythia-410M & 410M & 2.58 & 1245.00 & 4.92 \\
\hline
Mistral-Nemo-Base-2407   & 12B   & 2.82 & 1978.10 &  6.72  \\
\hline
Cerebras-GPT-1.3B         & 1.3B  & 2.86 & 1040.17 &  5.25  \\
\hline
Minitron-8B-Base            & 8B    & 2.98 & 2136.14 &  5.10  \\
\hline
Llama-3.1-8B            & 8B    & 3.22 &  970.64 &  6.38  \\
\hline
Opt-2.7B                  & 2.7B  & 3.59 & 1052.79 &  6.29  \\
\hline
Llama-3.2-3B            & 3B    & 3.74 & 1560.66 &  7.42  \\
\hline
Opt-1.3B                  & 1.3B  & 3.87 & 1121.87 &  6.85  \\
\hline
Phi-4                    & 14B   & 4.01 & 1163.23 & 12.07  \\
\hline
Llama-3.2-1B            & 1B    & 4.11 & 1718.67 &  8.57  \\
\hline
TinyLlama-1.1B-Chat-v1.0 & 1.1B    & 4.38 & 1014.17 &  8.95  \\
\hline
Stablelm-zephyr-3B     & 3B    & 4.57 & 2129.20 &  7.54  \\
\hline
Opt-350M                  & 350M  & 4.76 & 1075.87 &  8.19  \\
\hline
Mistral-7B-v0.3          & 7B    & 5.26 &  921.40 & 10.19  \\
\hline
LaMini-GPT-1.5B             & 1.5B    & 5.41 & 1322.01 &  9.49  \\
\hline
Hormoz-8B                   & 8B    & 5.98 & 4175.56 & 13.64  \\
\hline
Qwen2-7B                      & 7B    & 6.29 & 1356.97 & 14.17  \\
\hline
Bloom-3B              & 3B    & 6.95 & 3258.46 & 15.97  \\
\hline
SmolLM3-3B           & 3B    & 6.98 & 5372.60 & 13.95  \\
\hline
Qwen1.5-1.8B                  & 1.8B    & 8.07 & 1584.79 & 17.67  \\
\hline
Zephyr-7B-beta       & 7B    & 8.29 & 1515.70 & 15.82  \\
\hline
Bloom-1b1               & 1B    &11.01 & 3558.50 & 24.63  \\
\hline
Bloom-560M              & 560M  &12.64 & 4915.05 & 28.12  \\
\hline
Gemma-7B                    & 7B    &21.14 & 3152.45 & 66.52  \\
\hline
\end{tabular}
\label{ppt_table}
\end{table}

\subsection{Script Sensitivity Across Unicode, Romanized and Mixed-Script Text}

To quantify script sensitivity, model rankings and perplexity ratios across conditions were compared. Unicode performance does not translate to Romanized: Pythia-410M achieves Unicode perplexity 2.58 but Romanized 1,245.0, while Mistral-7B shows the reverse pattern (Unicode: 5.26, Romanized: 921.4). Romanized perplexity inflates by 1-2 orders of magnitude for most models, with BLOOM variants showing Unicode 6.95-12.64 but Romanized 3,200-4,900.

Mixed-script evaluation reveals distinct robustness patterns, with Pythia-410M, Minitron-8B and Cerebras-GPT-1.3B maintaining stable performance across all scripts. Conversely, some Unicode-strong models degrade sharply on mixed text (BLOOM variants: 13-28), indicating difficulty integrating heterogeneous scripts. Notably, Mistral-Nemo-Base-2407 (12B) achieves strong Unicode (2.82) and mixed (6.72) scores but struggles with pure Romanized (1978.1), demonstrating that mixed-script robustness differs from Romanized capability.

\subsection{Scaling Behavior and Family-Level Trends}

\begin{figure}[h]
\centering
\includegraphics[width=0.7\columnwidth]{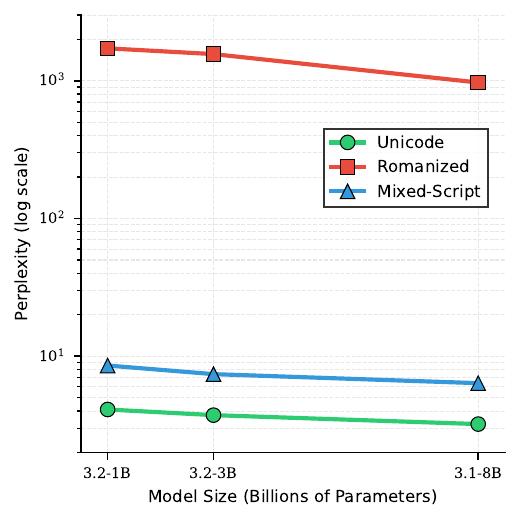}
\caption{Llama family scaling across script conditions (log scale).}
\label{fig:llama_scaling}
\end{figure}

The model suite allows examination of scaling trends within and across model families. Within the Llama family, performance improves consistently with scale across all script conditions, as illustrated in Fig.~\ref{fig:llama_scaling}. This suggests that, for well-designed architectures with appropriate pre-training data, increasing parameter count yields tangible gains in Sinhala modeling quality across scripts.

However, cross-family comparisons reveal that scaling is not sufficient on its own. Microsoft Phi-4 (14B) and Mistral-Nemo-Base-2407 (12B) achieve strong but not uniformly best results, with smaller models like Pythia-410M competing or outperforming them on Unicode and mixed-script text. Similarly, several 7B-8B models lag behind more compact models on specific script conditions, highlighting that architecture, tokenizer and data curation decisions can outweigh raw scale for low-resource, script-diverse languages.

A particularly notable pattern is that Llama-3.2-1B offers a favorable quality-efficiency trade-off among 1B-parameter models. It provides the best Unicode and mixed-script perplexities and competitive Romanized performance, making it a compelling choice for resource-constrained deployments. In contrast, some larger models exhibit unstable behavior, such as extremely high perplexity on at least one script condition, indicating limited suitability for robust multilingual deployment without further adaptation.

\subsection{Statistical Analysis of Script Sensitivity}

To quantify the degree of script-dependent performance divergence, Spearman rank correlations ($\rho$) were computed between model rankings across script conditions, with statistical significance assessed via p-values. The analysis reveals three key statistical patterns.

\begin{figure}[h]
\centering
\includegraphics[width=\columnwidth]{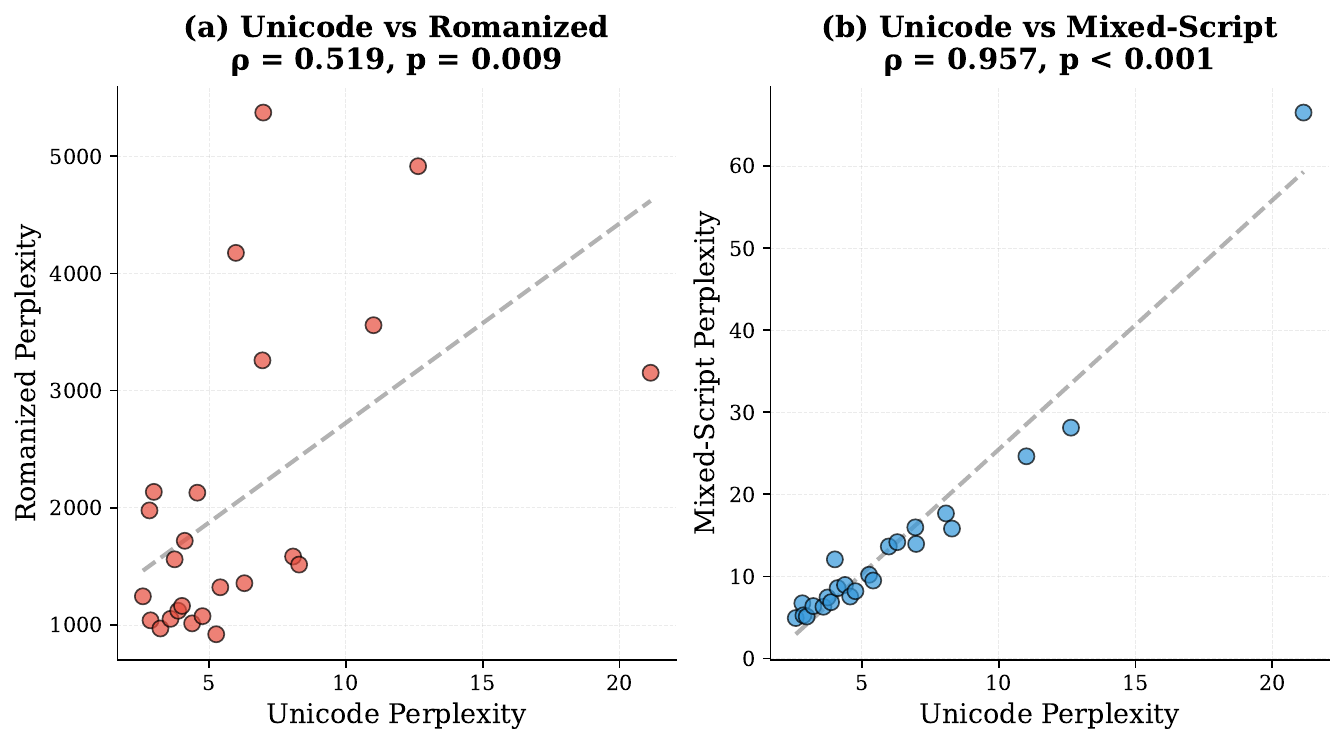}
\caption{Correlation between Unicode and other script conditions. (a) Moderate correlation with Romanized ($\rho$ = 0.519). (b) Strong correlation with mixed-script ($\rho$ = 0.957).}
\label{fig:correlation}
\end{figure}

First, Unicode and mixed-script rankings exhibit a very strong correlation ($\rho$ = 0.957, p $<$ 0.001), indicating that models with strong Unicode performance reliably maintain that advantage under script mixing. However, Unicode and Romanized rankings show only moderate correlation ($\rho$ = 0.519, p = 0.009), demonstrating that strong Unicode performance does not consistently predict Romanized capability. Similarly, Romanized and mixed-script rankings are moderately correlated ($\rho$ = 0.514, p = 0.010), suggesting that the two conditions measure somewhat distinct competencies. As illustrated in Fig.~\ref{fig:correlation}, the visual contrast between the scattered distribution in panel (a) and the tight linear pattern in panel (b) clearly demonstrates this divergence. The top five model overlap analysis confirms that only 2 of 5 top-performing Unicode models remain in the top five for Romanized text, whereas 4 of 5 maintain top five status for mixed-script evaluation.

Second, performance degradation from Unicode to Romanized script is severe. The median perplexity ratio is 312.3$\times$ (range: 149.1$\times$ to 769.7$\times$), with some models degrading by nearly three orders of magnitude. In contrast, mixed-script text shows modest degradation with a median ratio of 2.0$\times$ (range: 1.6$\times$ to 3.1$\times$), confirming that the presence of Unicode tokens substantially anchors model predictions even when Romanized segments are interspersed.

Third, the model parameter count shows negligible correlation with performance across all script conditions (Unicode: $\rho$ = 0.085, p = 0.693; Romanized: $\rho$ = -0.085, p = 0.693; Mixed-Script: $\rho$ = -0.018, p = 0.934). These near-zero correlations provide statistical evidence that model scale does not predict script-handling competence for low-resource languages, reinforcing the observation that architectural choices, tokenizer design and training data composition outweigh raw parameter count.

Variability in model performance, measured by coefficient of variation, was highest for mixed-script text (CV = 92.6\%) compared to Unicode (CV = 66.4\%) and Romanized (CV = 62.3\%), indicating that mixed-script evaluation exposes the widest range of model capabilities and robustness differences. These patterns suggest that existing multilingual benchmarks that evaluate Sinhala only in Unicode form substantially underestimate real-world model deficiencies, particularly for the majority of informal digital communication that occurs in Romanized or mixed-script form.

Together, these statistical patterns motivate a deeper examination of the mechanisms driving script sensitivity, explored in the following section.

\section{Discussion}

Current language models exhibit strong script sensitivity when modeling Sinhala. Unicode Sinhala is consistently easiest, with several models achieving perplexities below 4, yet the same architectures often fail on Romanized Sinhala, where perplexities inflate by one to two orders of magnitude, indicating poor representation in pre-training corpora or inadequate tokenization.

To illustrate, the Romanized sentence ``\textit{amma bath uyanawa}" (Mother is cooking rice) yields perplexity ratios of 836$\times$, 2723$\times$ and 6565$\times$ across Mistral-7B, Pythia-410M and Bloom-560M, respectively, despite \textit{bath} tokenizing as a single unit due to its overlap with English vocabulary. Genuinely Sinhala Romanized words such as \textit{uyanawa} $\rightarrow$ [uy, an, awa] and \textit{gewamu} $\rightarrow$ [ge, w, am, u] fragment consistently into meaningless subword units, confirming that the performance gap reflects the near-complete absence of Romanized Sinhala in pre-training corpora rather than tokenization granularity alone. These three models were selected to represent the full performance spectrum. Mistral-7B achieved the lowest Romanized perplexity (921.40) and Bloom-560M the highest (4915.05) among all 24 evaluated models, while Pythia-410M achieved the lowest Unicode perplexity (2.58), making it a natural reference point for cross-script comparison.

These qualitative observations are confirmed by statistical analysis. Spearman's rank correlation between Unicode and Romanized performance is only moderate ($\rho$ = 0.519, p = 0.009), meaning that nearly half of the ranking variance is script-dependent rather than reflecting general language modeling ability. The extreme perplexity degradation ratio (median 312$\times$) quantifies the severity of this divergence and suggests that Romanized Sinhala is effectively treated as an out-of-distribution condition by most models.

Mixed-script Sinhala further exposes robustness gaps that single-script evaluation cannot capture. While perplexities on mixed-script text generally fall between Unicode and Romanized values, model rankings shift noticeably. Models that are strong on Unicode sometimes degrade sharply under mixing, whereas a subset (notably Pythia-410M, Minitron-8B-Base and Cerebras-GPT-1.3B) maintain relatively stable performance across all three conditions. Robustness to script mixing thus depends on both overall language modeling strength and how well tokenizers handle heterogeneous script distributions.

Model scale shows nuanced effects. Within families such as Llama, larger models consistently improve perplexity across all scripts, indicating that scaling helps when architecture and data are well aligned with the language, as illustrated in Fig.~\ref{fig:llama_scaling}. However, cross-family comparisons reveal that small and medium-sized models can outperform much larger ones on Sinhala, particularly for mixed-script text, challenging the assumption that ``bigger is always better'' in low-resource, multi-script settings. The strong performance of Pythia-410M highlights the importance of targeted pre-training and tokenizer design over sheer parameter count.

From an application perspective, the divergence between Unicode and Romanized performance has direct implications for system design. Systems trained or evaluated solely on Unicode Sinhala risk catastrophic degradation when deployed in environments dominated by Romanized or mixed-script user input, such as social media, messaging platforms and online forums. Therefore, the results motivate script-aware evaluation pipelines and, where possible, preprocessing strategies such as robust transliteration, script normalization or targeted fine-tuning on Romanized and mixed-script data. The limitations are discussed in the following section.

\section{limitations}

The evaluation corpus of 1,500 sentences was constructed using K-means clustering to ensure semantic diversity, but may not fully capture dialectal variation or domain-specific usage. As the corpus is drawn primarily from social media platforms, it may over-represent informal registers while under-representing formal and domain-specific text, such as academic or professional writing.

Furthermore, perplexity as an intrinsic metric does not directly measure downstream task performance. The observed script sensitivity may manifest differently across specific applications such as machine translation, sentiment analysis or named entity recognition, and future evaluation should complement these findings with downstream task assessments to directly measure the practical impact of script sensitivity in real-world deployment scenarios.

\section{Conclusion}

This work presents the first systematic benchmark of language models on Unicode, Romanized and mixed-script Sinhala, revealing severe script sensitivity with 312× median performance degradation. Statistical analysis confirms model scale does not predict script-handling competence ($\rho$ $<$ 0.1), while Unicode performance strongly predicts mixed-script robustness ($\rho$ = 0.957) but not Romanized capability ($\rho$ = 0.519). These findings generalize to other low-resource languages with script duality, including Hindi, Tamil and Arabic dialects, where Romanized variants dominate digital communication despite limited model support. The evaluation framework introduced here can serve as a replicable protocol for benchmarking such script-diverse languages. 

The evaluation datasets are publicly available, enabling researchers working on other script-diverse low-resource languages to adopt this protocol directly for benchmarking language model performance across script conditions. Future work should also investigate the effect of cross-lingual vocabulary overlap on tokenization, since Romanized Sinhala words that coincidentally match English or other South Asian language words (e.g., \textit{bath}, \textit{api}) are tokenized more cleanly than genuinely unique Sinhala Romanized words, which may underestimate the true severity of tokenizer deficiency. Additionally, evaluating whether transliterating Romanized Sinhala to Unicode prior to inference can reduce the perplexity gap would offer a practical and immediately deployable solution for improving model performance on Romanized input without requiring retraining.

\end{document}